\documentclass[sigconf]{acmart}
\AtBeginDocument{%
  }

\copyrightyear{2026}
\acmYear{2026}
\setcopyright{cc}
\setcctype{by}
\acmConference[MM '26]{Proceedings of the 34th ACM International Conference on Multimedia}{November 10--14, 2026}{Rio de Janeiro, Brazil}
\acmBooktitle{Proceedings of the 34th ACM International Conference on Multimedia (MM '26), November 10--14, 2026, Rio de Janeiro, Brazil}
\acmDOI{10.1145/3767308.3835295}
\acmISBN{979-8-4007-2213-4/2026/11}

\usepackage[table]{xcolor}
\begin{document}

\title{Preserve More Details: Mitigating Content Drift in Real-World Image Super-Resolution}

\author{Chunxiao Liu}
\orcid{0009-0007-0275-7567}
\affiliation{%
  \institution{Xiaomi Corporation}
  \city{Beijing}
  \country{China}
}
\email{liuchunxiao@xiaomi.com}

\author{Wei Liu}
\authornote{Corresponding author.}
\orcid{0009-0003-6935-2231}
\affiliation{%
  \institution{Xiaomi Corporation}
  \city{Beijing}
  \country{China}
}
\email{liuwei67@xiaomi.com}

\author{Anbin Xiong}
\orcid{0009-0009-7411-0003}
\affiliation{%
  \institution{Xiaomi Corporation}
  \city{Beijing}
  \country{China}
}
\email{xionganbin@xiaomi.com}

\author{Erli Meng}
\orcid{0009-0009-3748-126X}
\affiliation{%
  \institution{Xiaomi Corporation}
  \city{Beijing}
  \country{China}
}
\email{mengerli@xiaomi.com}


\begin{abstract}
 Real-world image super-resolution (Real-ISR) aims to reconstruct high-quality (HQ) images from low-quality (LQ) inputs subject to diverse real-world degradations. Recent advances have leveraged the LQ inputs and natural image priors learned by Stable Diffusion models to achieve impressive results. However, existing methods often overlook insufficient clarity of LQ inputs inevitably induce content drift in the generated HQ images. This manifests primarily as visual detail degradation and textual semantic shift, severely compromising both fidelity and perceptual quality. To address this challenge, we propose FSP-Diff, a novel one-step diffusion model featuring a dual-pathway architecture. This architecture comprises a Detail-Conditioned Pathway for injecting structured details to recover fine structures, and a Detail-Modulated Semantic Pathway that refines semantic guidance using structured details to mitigate semantic deviations. Extensive experiments on standard Real-ISR benchmarks demonstrate that FSP-Diff surpasses existing one-step diffusion methods in both quantitative and qualitative metrics. 
\end{abstract}

\ccsdesc[500]{Computing methodologies~Reconstruction}

\keywords{Real-world image super-resolution, One-step diffusion, Multimodal, Diffusion model}


\maketitle
\typeout{ACMART VERSION >>> \csname ver@acmart.cls\endcsname}

\section{Introduction}
\label{sec:intro}

Real-world image super-resolution (Real-ISR) aims to reconstruct high-quality (HQ) images from low-quality (LQ) inputs captured in unconstrained scenarios \cite{zhang2021designing,wang2021real}. Unlike conventional SR tasks that assume synthetic and predefined degradations \cite{lim2017enhanced,chen2021pre,zhang2022efficient,dai2019second,liang2021swinir,zhang2018image}, Real-ISR requires handling complex, unknown, and spatially variant real-world degradations. This inherent uncertainty demands models that not only recover fine structures, but also generate perceptually realistic results aligned with the natural image.

The prevalent paradigm for Real-ISR training involves synthesizing LQ-HQ pairs via sophisticated degradation pipelines \cite{zhang2021designing,wang2021real}, enabling large-scale learning without expensive real-world annotations. Based on such synthesized training data, text-conditioned Stable Diffusion (SD) models \cite{rombach2022high,saharia2022photorealistic,esser2024scaling} have emerged as a powerful framework for this task. This is due to the fact that SD-based methods \cite{lin2024diffbir,wang2024exploiting,wu2024seesr,yang2024pixel} inject strong natural image priors learned from billions of image-text pairs while pre-training. By seamlessly integrating these priors into the super-resolution process, these approaches effectively enhance the perceptual naturalness of the resulting super-resolution outputs.
\begin{figure}[tb]
  \centering
  \includegraphics[trim=0 10 0 0, clip, width=\linewidth]{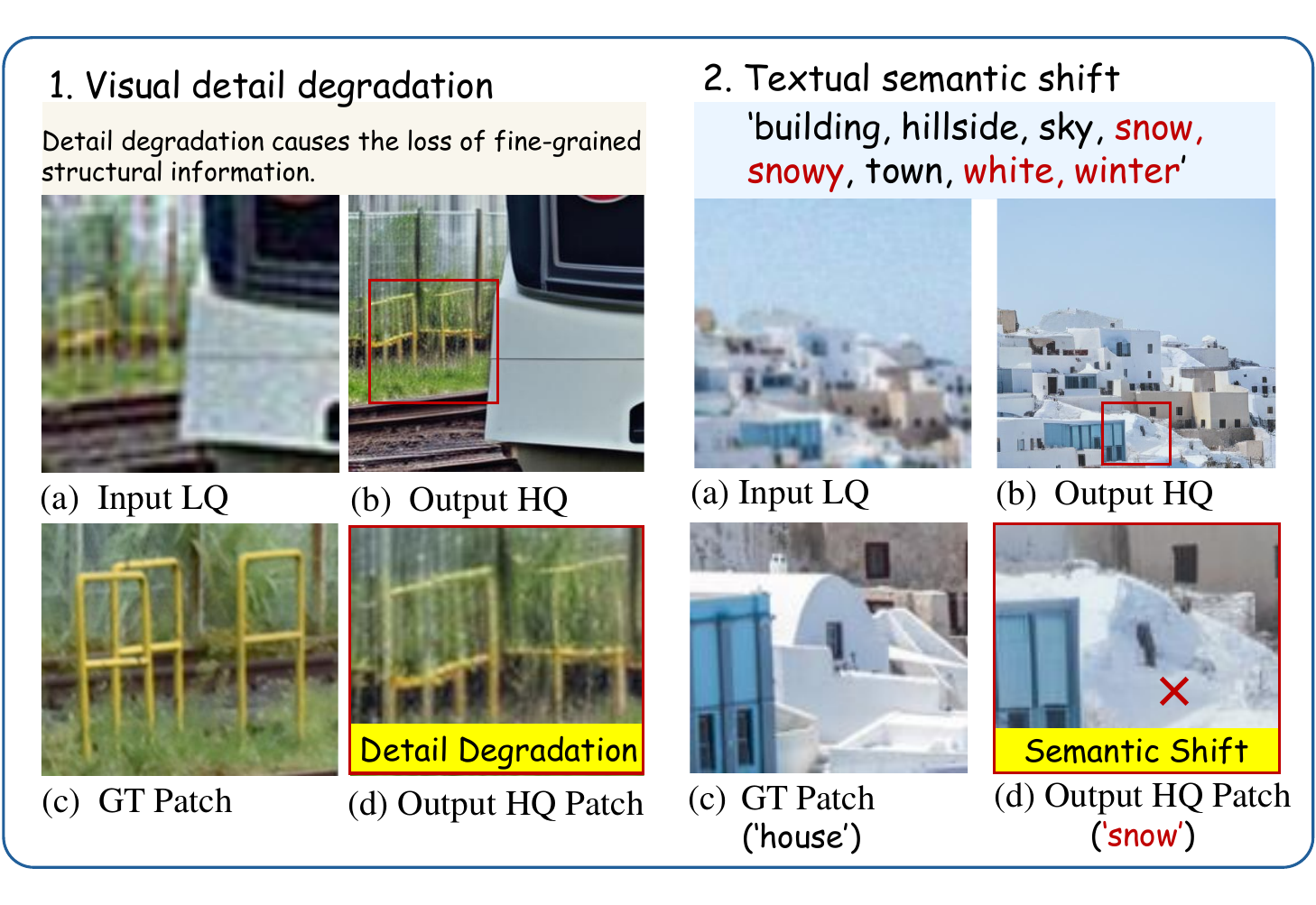}
  \caption{Content drift in Real-World Image Super-Resolution is highlighted by red boxes. The left panel illustrates visual detail degradation caused by distorted local structures. The right panel demonstrates a textual semantic shift, where ``$\textcolor{red}{\times}$'' marks incorrect semantic reconstructions shifting from ``house'' to ``snow''.}
  \label{fig:real-sr_intro}
\end{figure}

However, existing approaches overlook a critical limitation: insufficient clarity of LQ images induces content drift in generated HQ outputs. This drift manifests in two aspects: visual detail degradation and textual semantic shift, both degrading fidelity and perceptual quality. First, the highly compressed latent space of the VAE in Stable Diffusion \cite{rombach2022high} discards fine structures of LQ inputs, making the HQ generation process overly reliant on pre-trained priors rather than LQ intrinsic content. This causes degraded visual details in HQ outputs, especially for delicate elements (e.g., fences shown in the left panel of Fig.~\ref{fig:real-sr_intro}. Second, existing methods typically generate captions from LQ images as semantic guidance for HQ synthesis, but LQ images with unclear details induce textual semantic shift. As illustrated in the right panel of Fig.~\ref{fig:real-sr_intro}, this can lead to severe misreconstruction (e.g., a house incorrectly rendered as a snow pile). Thus, how to mitigate content drift by preserving fine structures of LQ image is an essential problem for real-world super-resolution.

To address the aforementioned issue, we propose FSP-Diff, a Fine Structure Preserving diffusion model that mitigates the content drift between LQ and HQ image by preserving fine structures and providing robust semantic guidance. Specifically, FSP-Diff introduces a dual-pathway architecture, including a \textbf{Detail-Conditioned Pathway} and a \textbf{Detail-Modulated Semantic Pathway}, which act as detail injector and semantic inspector, respectively. The detail injector enables precise recovery of fine structures by injecting structured details using structured detail conditioned attention. The semantic inspector adaptively refines semantic guidance based on structured details. Such a dual-pathway design enables the diffusion model to effectively suppress visual detail degradation and textual semantic shift induced by unreliable LQ inputs. Extensive experiments conducted on standard benchmarks ~\cite{wei2020component,cai2019toward,agustsson2017ntire} demonstrate that FSP-Diff achieves significant performance improvements, consistently outperforming existing one-step diffusion-based methods in both quantitative metrics and qualitative visual quality. Our contributions can be summarized as:

(1) We propose that unreliable LQ inputs inevitably lead to content drift of generated HQ images, causing visual detail degradation and textual semantic shift.

(2) To mitigate the content drift, we propose a novel dual-pathway architecture that injects structured details in the Detail-Conditioned Pathway, and suppresses the unreliable semantic in the Detail-Modulated Semantic Pathway. These pathways jointly improve the fidelity and reliability of real-world image super-resolution.

(3) Extensive experiments on standard benchmarks show that our approach can achieve significant improvement just in one-step, preserving more structured details in generated HQ images.

\section{Related Work}

\subsection{GAN-based Real-ISR methods}
GAN-based frameworks~\cite{chen2022real,liang2021swinir,liang2022details,liang2022efficient,xie2023desra,zhang2021designing,wang2021real} have been actively explored to simulate real-world degradations and enhance the photorealism of ISR results. For instance, BSRGAN~\cite{zhang2021designing} and Real-ESRGAN~\cite{wang2021real} generate LQ–HQ pairs using randomized and high-order degradation pipelines, respectively, providing more realistic training data and inspiring subsequent works. DASR~\cite{liang2022efficient} introduces a degradation-adaptive network that dynamically adjusts its parameters based on the estimated degradation of each input image. LDL~\cite{liang2022details} uses local statistics to suppress artifacts and enhance perceptual quality.
Despite these advances, training GANs~\cite{goodfellow2014generative} remains unstable, and the discriminator is limited in evaluating diverse natural image content, often resulting in unnatural visual artifacts in the reconstructed images.
\subsection{Diffusion-based Real-ISR methods}
Diffusion models~\cite{ho2020denoising,song2020denoising,dhariwal2021diffusion,song2020score} have recently gained increasing attention in Real-ISR, driven by their strong generative capabilities. Some works~\cite{yue2023resshift,kawar2022denoising} train diffusion models from scratch, with their performance limited by the scale and diversity of existing datasets. Consequently, pre-trained text-to-image diffusion models~\cite{saharia2022photorealistic,rombach2022high,podell2023sdxl,chen2023pixart,zhang2023adding} are commonly adapted by current Real-ISR methods~\cite{lin2024diffbir,qu2024xpsr,wang2024exploiting,sun2025pixel,xie2024addsr,wu2024one}, typically by encoding LQ images into a compact latent space for efficiency, followed by latent-conditioned diffusion to reconstruct high-resolution details. In practice, diffusion-based Real-ISR methods generally fall into two categories: multi-step and one-step approaches.

\noindent\textbf{Multi-step Diffusion-based Real-ISR.}
Starting from an initial state, multi-step diffusion frameworks~\cite{yang2024pixel,yu2024scaling,wang2024exploiting,chen2025faithdiff,wu2024seesr} iteratively refine the sample through a diffusion process to obtain an HQ image. StableSR~\cite{wang2024exploiting} exploits pre-trained text-to-image(T2I) diffusion priors with a time-aware encoder and employs controllable feature aggregation to balance fidelity and perceptual quality. SeeSR~\cite{wu2024seesr} introduces a semantics-aware diffusion framework that employs degradation-robust soft and hard prompts to preserve semantic fidelity during super-resolution. Considering the gap between degraded inputs and diffusion latents, FaithDiff~\cite{chen2025faithdiff} leverages latent diffusion priors with feature alignment and  the strong captioning
capability of LLAVA~\cite{liu2023visual} to recover structurally consistent HQ images. While multi-step diffusion methods achieve promising Real-ISR performance, their reliance on numerous diffusion steps (often ten or more) leads to high computational cost and latency.

\noindent\textbf{One-step Diffusion-based Real-ISR.}
Recently, several one-step diffusion-based Real-ISR methods~\cite{zhang2024degradation,luo2023latent,wang2025osdface} have been proposed to address the inference inefficiency of multi-step diffusion. SinSR~\cite{wang2024sinsr} distills ResShift~\cite{yue2023resshift} into a single-step SR model using consistency-preserving loss for efficient high-resolution reconstruction, but its generalization remains constrained by limited training data. OSEDiff~\cite{wu2024one} condenses a pre-trained multi-step T2I diffusion model into a one-step Real-ISR framework, it directly takes the LQ image as input, leveraging lightweight LoRA adapters~\cite{hu2022lora} and latent-space variational score distillation (VSD)~\cite{wang2023prolificdreamer,yin2024one,dao2024swiftbrush} for efficient training. Building on this, TSD-SR~\cite{dong2025tsd} introduces target score distillation and a distribution-aware sampling module to improve gradient reliability and  guide recovery toward a high-fidelity distribution, enabling more effective one-step super-resolution. 

Our proposed FSP-Diff adopts a one-step diffusion paradigm that maps LQ images directly to HQ outputs. Unlike existing one-step methods such as OSEDiff~\cite{wu2024one}, FSP-Diff explicitly extracts and preserves more structured detail, effectively leveraging them for HQ image generation. Among recent efforts, TVT~\cite{yi2025fine} improves the reconstruction of fine structures by transferring a heavily compressed VAE (originally trained at  8× compression) to a less aggressive 4× variant through staged encoder–decoder adaptation. However, this approach requires large-scale VAE retraining on the OpenImages dataset~\cite{kuznetsova2020open} and involves a complex multi-stage training procedure.

\section{Method}
In this section, we first review the general paradigm of diffusion-based models for Real-ISR, followed by the motivation underlying our design. Then we present the proposed FSP-Diff framework along with its key components, and finally detail the model optimization strategies.

\subsection{Problem Formulation}
\label{sec:blind}

Given an LQ image $I_L $, existing one-step diffusion-based approaches~\cite{wu2024one,dong2025tsd,yi2025fine} generally adopt a deterministic diffusion formulation~\cite{wang2024sinsr} to establish the LQ-to-HQ mapping. Specifically, $I_L $ is encoded into a latent space to obtain its representation $Z_L$,  which serves as the starting point of the generation process and is directly fed into a pre-trained T2I diffusion model~\cite{rombach2022high} without noise injection, producing a deterministic HQ output in a single step. The formulation is given as follows:
\begin{equation}
\hat{I}_H = G_\theta(Z_L, C_T)
\end{equation}
where $G_{\theta}$ denotes the generative model parameterized by $\theta$, $\hat{I}_H$ is the generated HQ image and ${C_T}$ is the textual guidance. 

Following the above framework, promising LQ-to-HQ generation results can generally be achieved. However, the compact latent-space compression inevitably discards part of structured detail information. As a result, the generated images may exhibit visual detail degradation
and textual semantic shift. Our work focuses on mitigating these two limitations, we first extract a structured details $X_S$ which preserves richer fine structures compared to the highly compressed latent $Z_L$. Based on this, we introduce two enhancement pathways with learnable components for detail injection and semantic refinement, denoted as $P_t$ and $P_s$ respectively. Therefore, the overall generation process is further reformulated as:
\begin{equation}
\hat{I}_H = G_\theta(Z_L, P_t(X_S), P_s(C_T,X_S))
\end{equation}

We train $G_{\theta}$ on a dataset $\mathcal{D}$ of $(I_L, I_H)$ pairs with a reconstruction loss $\mathcal{L}_{\mathrm{rec}} $ and a VSD-based regularization loss $\mathcal{L}_{\mathrm{reg}}$~\cite{wu2024one}. Here, $\mathcal{L}_{\mathrm{rec}} $ encourages $\hat{I}_H$ to be close to the ground-truth image $I_H$, commonly instantiated with distance-based metrics such as e.g., $L_1$ norm, $L_2$ norm, or LPIPS~\cite{zhang2018unreasonable}, while $\mathcal{L}_{\mathrm{reg}}$ tends to improve the generalization capability by aligning the outputs with the distribution of real-world HQ images, the formulation is:
\begin{equation}
\theta^* = \arg\min_\theta \mathbb{E}_{(I_L, I_H) \sim \mathcal{D}} \left[ \mathcal{L}_{\text{rec}}\left(G_\theta(I_L), I_H\right) + \lambda \mathcal{L}_{\text{reg}}\left(G_\theta(I_L)\right) \right]
\end{equation}
where $\lambda$ is a weighting factor balancing the training objectives.
\subsection{Overview of FSP-Diff}
\begin{figure*}[tb]
  \centering
  \includegraphics[width=0.8\linewidth]{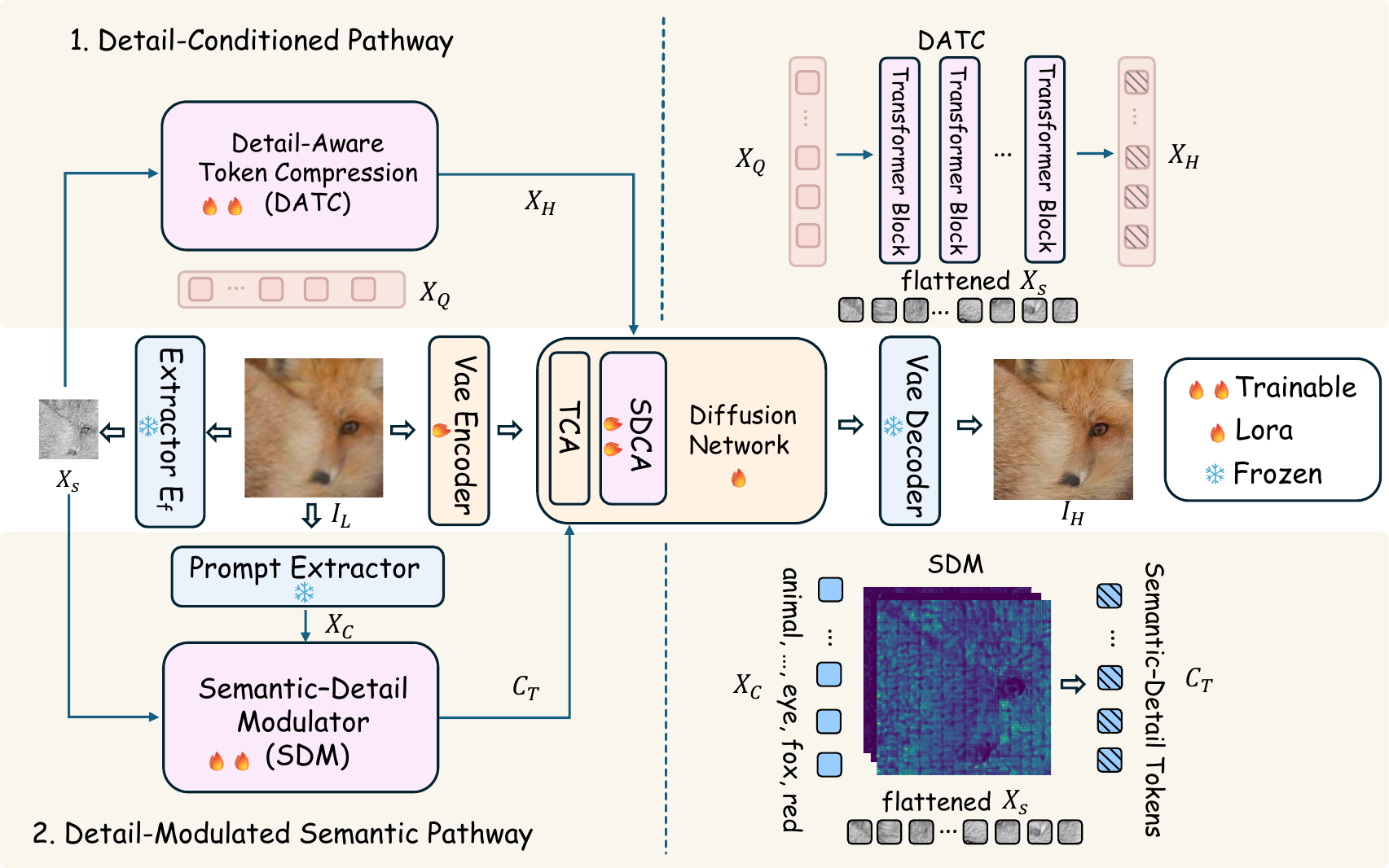}
  \caption{Overview of our method. We first extract a structured details $X_S$ and construct two complementary pathways. The former aggregates fine structures via a Detail-Aware Token Compression module (DATC),  followed by a Structured Detail Conditioned Attention module (SDCA) to guide detailed image reconstruction, while the latter adaptively modulates textual semantics using a Semantic–Detail Modulator(SDM) conditioned on the structured details.
  }
  \label{fig:real-sr framework}
\end{figure*}
As depicted in Fig.~\ref{fig:real-sr framework}, our model builds upon a pre-trained T2I generative framework, which comprises a VAE encoder $E_{\theta}$ for latent-space compression, a diffusion network $\varepsilon_{\theta}$ for LQ-to-HQ distribution modeling, and the corresponding VAE decoder $D_{\theta}$. Following the design of OSEDiff~\cite{wu2024one}, we fine-tune the diffusion model and VAE encoder via LoRA ~\cite{hu2022lora} while keeping the VAE decoder frozen. 

Based on this, we propose FSP-Diff, which casts LQ-to-HQ generation as a one-step diffusion process with two pathways guided by structured details: a Detail-Conditioned Pathway and a Detail-Modulated Semantic Pathway, corresponding to $P_t$ and $P_s$ described in Section~\ref{sec:blind}, respectively. The former explicitly extracts, transforms, and integrates high-frequency structured details to preserve fine-grained details during generation, while the latter jointly exploits textual tokens and structured details. The Detail-Modulated Semantic Pathway leverages the structured details to adaptively rectify potential semantic misalignment in the textual tokens, facilitating more reliable semantic conditioning during HQ generation.

\subsection{Detail-Conditioned Pathway}
\label{sec:tcp}
\noindent\textbf{Structured Details Extraction.}
To obtain image features with richer structured details, we use a pre-trained Vision Transformer (ViT) \cite{dosovitskiy2020image} as backbone, and adopt the first block of ViT encoder as our feature extractor $E_f$, since its early layers mainly capture low-level edge and boundary information. The parameters of $E_f$ are initialized from the ``$vit\_b$''model and kept frozen during training to preserve the pre-trained representations. Given a low-quality image  $I_L \in \mathbb{R}^{H \times W \times C}$, $E_f$ extracts structured detail with a spatial resolution of $\frac{H}{8}\times \frac{W}{8}$, denoted by $X_{S}$, which serve as the initial representation of structured details, providing informative guidance for subsequent detail enhancement.

\noindent\textbf{Detail-Aware Token Compression.}
After obtaining the spatial feature map $X_{S}$ from the extractor $E_f$, we explore integrating it into the diffusion generation process. A straightforward approach is to flatten it into $L_s = \frac{H}{8}\times \frac{W}{8}$ tokens, each with dimension $d$.  However, the resulting long token sequence makes the attention operations in subsequent modules computationally expensive. When interacting with the latent feature map $Z_L$ of size $\frac{H}{8}\times \frac{W}{8}$, the computational cost grows quadratically with the sequence length, i.e., $\mathcal{O}(N^2)$. To mitigate this, we propose a Detail-Aware Token Compression (DATC) module:

1) Initialize $L_q$ learnable query vectors $X_Q=\{q_1,q_2,…,q_{L_q}\}^\top \in \mathbb{R}^{L_q \times d}$, where $ L_q \ll L_z$ and $ L_q \ll L_s$, and $L_z$ denotes the sequence length of the latent feature $Z_L$.

2) Build a $M$-layer Transformer~\cite{vaswani2017attention}, where each layer first applies self-attention to model global context, followed by cross-attention using updated queries to selectively aggregate structured detail from $X_{S}$.

By stacking $M$ Transformer blocks, $X_{S}$ is progressively refined into a compact yet informative token set $X_H=\{h_1,h_2,…,h_{L_q}\}^\top \in \mathbb{R}^{L_q \times d}$, preserving fine structures while reducing the sequence length to a tractable scale suitable for efficient computation.

\noindent\textbf{Structured Detail Conditioned Attention.}
Building on $X_H$, we design a Structured Detail Conditioned Attention (SDCA) module that explicitly guides the diffusion-based generation process. Concretely, the SDCA module implements an attention mechanism in which the diffusion latents $Z_L$ serve as queries, and the structured details $X_H$ serves as keys and values. The SDCA modules are integrated into each UNet block of Diffusion model $\varepsilon_{\theta}$ and are inserted after text cross-attention (TCA) module, enabling explicit aggregation of structured details to guide fine-structure HQ generation. We can formulate this process as:
\begin{equation}
Z_L^{(i+1)} = \mathrm{SDCA}\left( \mathrm{TCA}\left( Z_L^{(i)}, C_T \right), X_H \right)
\end{equation}
where $ Z_L^{(i)} \in \mathbb{R}^{L_z \times d} $ denotes the input latent to the $i$-th UNet, and $C_T$ is the textual guidance defined in Section \ref{sec:tms}.

\subsection{Detail-Modulated Semantic Pathway }
\label{sec:tms}
To leverage T2I diffusion priors for semantic guidance, we first adopt the DAPE module~\cite{wu2024seesr} to generate text prompts describing the input image, and subsequently use the CLIP text encoder~\cite{rombach2022high} to obtain the corresponding semantic embeddings $X_C \in \mathbb{R}^{L_c \times d}$. Then we introduce the Semantic–Detail Modulator (SDM) to adaptively refine the original textual semantics by leveraging structured details $X_S$. Specifically, SDM employs an attention-based interaction mechanism: given $X_S$ and the semantic embeddings $X_C$, the semantic–detail correlation is computed via scaled dot-product attention, the attention weights are obtained as:
\begin{equation}
A = \mathrm{softmax}\!\left(\frac{X_C W_Q (X_S W_K)^\top}{\sqrt{d}}\right)
\end{equation}
The detail-aware modulation signal is then aggregated as:
\begin{equation}
C_T = A \cdot (X_S W_V)
\end{equation}
where $W_Q$ , $W_K$ and $W_V$ denote learnable projection matrices. Finally, $C_T$ supersedes the original text encoding $X_C$ and employed as the conditioning text unit to guide the diffusion model during generation.

\subsection{Loss Formulation and Timestep-Shifted Training}
As discussed in Section~\ref{sec:blind}, effective training of a Real-ISR model requires the generator to be supervised by both a data fidelity term $\mathcal{L}_{\mathrm{rec}}$  and a regularization term $\mathcal{L}_{\mathrm{reg}}$. We optimize the generator $G_{\theta}$ by:
\begin{equation}
\mathcal{L}\left(G_\theta(I_L), I_H\right) = \mathcal{L}_{\text{rec}}\left(G_\theta(I_L), I_H\right) + \lambda \mathcal{L}_{\text{reg}}\left(G_\theta(I_L)\right)
\end{equation}
As for $\mathcal{L}_{\mathrm{rec}}$, we adopt the weighted sum of $L_1$ loss and LPIPS loss:
\begin{equation}
\mathcal{L}_{\text{rec}}\left(G_\theta(I_L), I_H\right) = \mathcal{L}_{L_1}\left(G_\theta(I_L), I_H\right) + \lambda_1 \mathcal{L}_{\text{LPIPS}}\left(G_\theta(I_L), I_H\right)
\end{equation}
As in OSEDiff, we apply two regularizers in the latent space: a frozen pre-trained SD model $\varepsilon_{\phi}$ and its LoRA-finetuned replica $\varepsilon_{\phi'}$. The generator $G_{\theta}$ can be regulated via:
\begin{equation}
\mathcal{L}_{\text{reg}}(G_\theta(I_L)) = \mathcal{L}_{\text{VSD}}(G_\theta(I_L), X_C)
\end{equation}
Given the latent output $\hat{Z}$ of the generator $G_{\theta}$ and the semantic embeddings $X_C$, the gradient with respect to $\theta$ is computed as:
\begin{equation}
\begin{aligned}
\nabla_\theta \mathcal{L}_{\text{VSD}}(\hat{Z}_H, X_C) = \mathbb{E}_{t,\sigma \sim \mathcal{N}(0, I)} \Big[ \omega(t) \Big( \varepsilon_{\phi}(\hat{Z}_t; t, X_C) \\
- \varepsilon_{\phi'}(\hat{Z}_t; t, X_C) \Big) \frac{\partial \hat{Z}}{\partial \theta}\Big]
\end{aligned}
\end{equation}
where $\hat{Z}_t = \alpha_t \hat{Z}_H + \beta_t \sigma$ ,$ \alpha_t$ and $\beta_t$ are the noise-data scaling constants. $\omega(t)$ is a time-varying weighting function detailed in ~\cite{wu2024one}, and $t \in \{1, \dots, T\}$ indicates that the gradient is computed in expectation over all diffusion timesteps.

Based on empirical exploration, we apply a shift to the timestep t sampling distribution prior to its application in $\hat{Z}_H$, where a scaling factor s controls whether t is biased toward lower (s<1) or higher (s>1) noise regime.
\begin{equation}
\text{shift}(t, s) := \frac{s \times t}{1 + (s - 1) \times t}
\end{equation}
The formulation follows \cite{lin2025diffusion}. In practice, we set $\lambda=1$, $\lambda_1=2$ and $s=0.9$ by default unless stated otherwise.

\section{Experiments}
We evaluate FSP-Diff by comparing it with current state-of-the-art one-step diffusion-based models for Real-ISR across multiple public benchmarks, covering metrics that assess both fidelity and perceptual quality. Furthermore, we conduct ablation studies to assess the contribution of each proposed component and the impact of key hyperparameters. Finally, we analyze the time complexity of our model.
\subsection{Experimental Settings}
\noindent\textbf{Training and Testing Datasets.}
To ensure consistency with prior work~\cite{wu2024one,sun2025pixel,dong2025tsd}, we evaluate the proposed method on the same datasets and adopt identical data preprocessing procedures, thereby facilitating a fair comparison with existing methods. Specifically, LSDIR dataset~\cite{li2023lsdir} and the first 10K face images from FFHQ~\cite{karras2019style} are used for training pairs, and the corresponding LQ images are generated using the Real-ESRGAN degradation pipeline~\cite{wang2021real}, forming the LQ–HQ training pairs. We evaluate our model and compare it with the state-of-the-arts on a test set comprising DIV2K-val~\cite{agustsson2017ntire}, RealSR~\cite{cai2019toward}, and DRealSR~\cite{wei2020component}. The DIV2K-val dataset contains 3,000 LR–HR image pairs with a resolution of 512 $\times$ 512, synthesized using the Real-ESRGAN degradation pipeline. RealSR and DRealSR are real-world datasets collected using long-short camera focal lenses, where the LQ and HQ images are of resolutions 128 $\times$ 128 and 512 $\times$ 512, respectively.

\noindent\textbf{Evaluation Metrics.}
To comprehensively evaluate our model, we utilize both reference-based and no-reference metrics. 
The reference-based metrics include PSNR and SSIM~\cite{wang2004image} computed on the Y channel in YCbCr color space for fidelity; LPIPS\cite{zhang2018unreasonable} and DISTS~\cite{ding2020image}, which evaluate perceptual similarity; and FID~\cite{heusel2017gans}, assessing the distributional similarity between ground-truth HQ images and restored outputs. The no-reference image quality metrics include NIQE~\cite{zhang2015feature}, which estimates naturalness; MANIQA~\cite{yang2022maniqa}, MUSIQ~\cite{ke2021musiq} and CLIPIQA~\cite{wang2023exploring}, which reflect perceptual quality and semantic consistency without relying on the ground-truth images.

\noindent\textbf{Compared Methods.}
We compare FSP-Diff with advanced SD-based Real-ISR methods, including SinSR~\cite{wang2024sinsr}, OSEDiff~\cite{wu2024one}, TSD-SR~\cite{dong2025tsd}, PiSA-SR~\cite{sun2025pixel}, and TVT~\cite{yi2025fine}, which are representative state-of-the-art one-step diffusion approaches. 
Notably, TVT adopts a multi-stage training strategy, first retraining its VAE on the OpenImages and LSDIR datasets, and then further fine-tuning its generative model on RealCE~\cite{ma2023benchmark} for real-world scene text image super-resolution. TSD-SR leverages a larger training dataset comprising DIV2K and Flickr2K~\cite{timofte2017ntire}, and employs a stronger generative model SD3~\cite{esser2024scaling} as its backbone. By contrast, our experiments strictly follow the setup of OSEDiff in terms of training data and generative backbone. No additional external scene datasets or more powerful generative foundation models are used, and the model is trained in a single stage.

\noindent\textbf{Implementation Details.}
We initialize our network from the pre-trained SD 2.1-base model. During training, we update all modules except for: (1) the structured detail extractor $E_f$, (2) the DAPE module and the ClIP text encoder used for image captions generation and embedding, and (3) the regularizer $\varepsilon_{\phi}$. The parameters of all three described above components are frozen.
We adopt the same VSD parameter update strategy as in OSEDiff~\cite{wu2024one}, and the rank of LoRA~\cite{hu2022lora} is set to 4 for the VAE encoder, the diffusion network, and the fine-tuned regularizer $\varepsilon_{\phi'}$.
We adopt the AdamW optimizer with a learning rate of 5e-5. The entire training process took approximately four days on a single NVIDIA H200 GPU with a batch size of 16.

\subsection{Comparison with State-of-the-Arts}
\begin{figure*}[tb]
  \centering
  \includegraphics[trim=0 61 0 50, clip, width=0.9\linewidth]{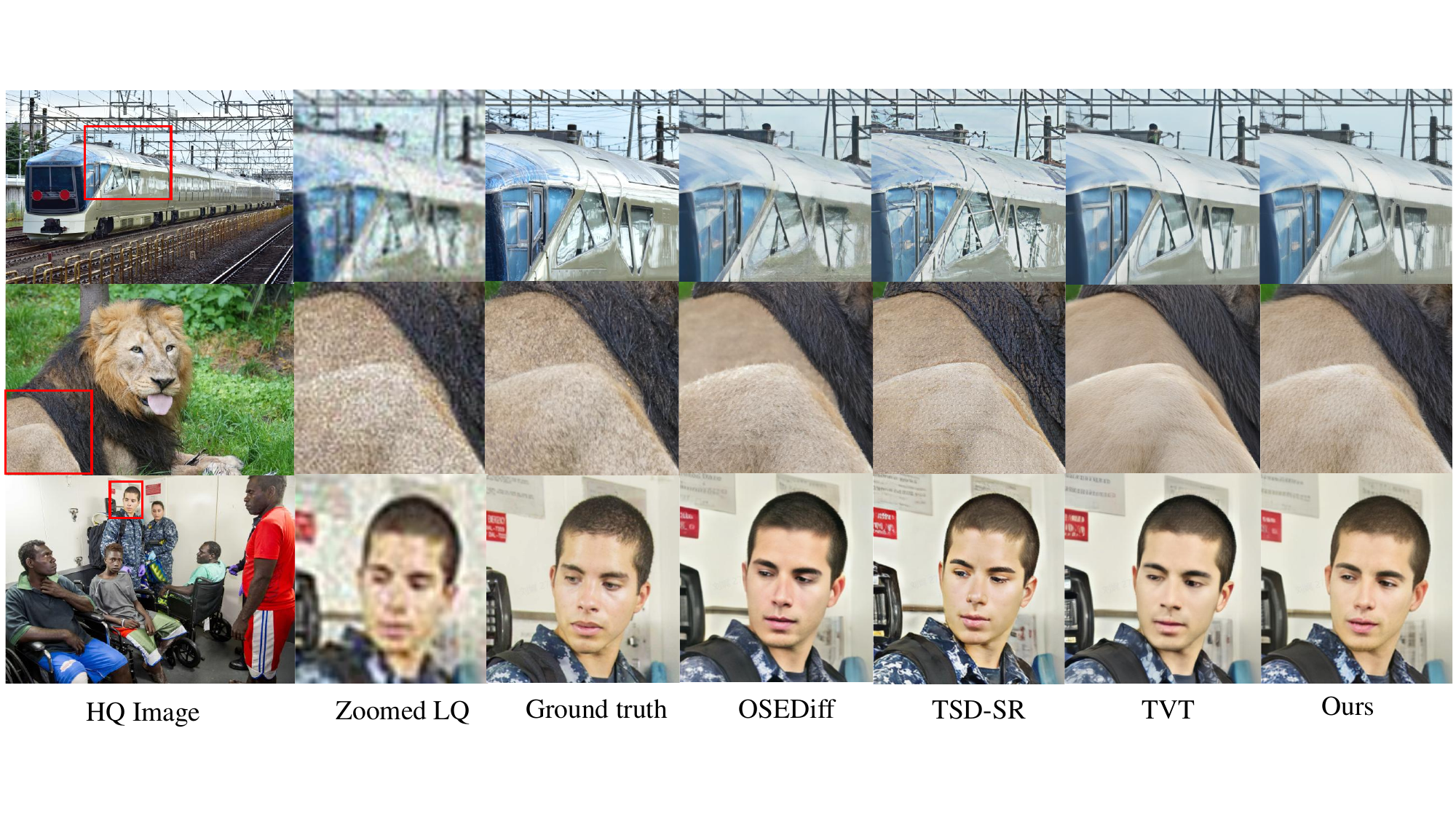}
  \caption{Visual Comparisions of different Real-ISR methods. Please zoom in for a better view.
  }
  \label{fig:sota_comp}
\end{figure*}
\begin{table*}[]
\caption{Quantitative comparison with state-of-the-art methods on both real-world and synthetic benchmarks.The best and second best results of each metric are highlighted in \textcolor[HTML]{D83931}{red} and \textcolor[HTML]{245BDB}{blue}, respectively. The symbol `\_' indicates that the corresponding metric outperforms that of OSEDiff.}
\label{tab:quantity}
\centering
\resizebox{\linewidth}{!}{
\begin{tabular}{l|l|lllllllll}
\hline
Datasets & Methods & PSNR ↑ & SSIM↑ & LPIPS ↓ & DISTS↓ & CLIPIQA↑ & NIQE↓ & MUSIQ↑ & MANIQA↑ & FID↓ \\ \hline
                      & SinSR~\cite{wang2024sinsr} & {\color[HTML]{245BDB} 28.41} & 0.7495 & 0.3741 & 0.2488 & 0.6367 & 7.02 & 55.34 & 0.4898 & 177.05 \\
                      & OSEDiff~\cite{wu2024one} & 27.92 & 0.7835 & 0.2968 & 0.2165 & 0.6963 & 6.49 & 64.65 & 0.5899 & 135.30 \\
                      & TVT~\cite{yi2025fine} & 28.25 & {\color[HTML]{245BDB} 0.7898} & {\color[HTML]{245BDB} 0.2900} & 0.2206 & {\color[HTML]{245BDB} 0.7227} & 7.03 & 65.60 & 0.5777 & 134.25 \\
         {DrealSR}    & TSD-SR~\cite{dong2025tsd} & 27.77 & 0.7559 & 0.2967 & {\color[HTML]{245BDB} 0.2136} & {\color[HTML]{D83931} 0.7344} & {\color[HTML]{D83931} 5.91} & {\color[HTML]{D83931} 66.62} & 0.5874 & 134.98 \\
                       & PiSA-SR~\cite{sun2025pixel} & 28.31 & 0.7804 & 0.2960 & 0.2169 & 0.6970 & {\color[HTML]{245BDB} 6.20} & {\color[HTML]{245BDB} 66.11} & {\color[HTML]{D83931} 0.6156} & {\color[HTML]{245BDB} 130.61} \\
                       & FSP-Diff & {\color[HTML]{D83931} {\underline{28.85}}} & {\color[HTML]{D83931} {\underline{0.7933}}} & {\color[HTML]{D83931} {\underline{0.2741}}} & {\color[HTML]{D83931} {\underline{0.2014}}} & 0.6868 & {\underline{6.24}} & {\underline{65.17}} & {\color[HTML]{245BDB} {\underline{0.6092}}} & {\color[HTML]{D83931} {\underline{127.19}}} \\ \hline
                       & SinSR~\cite{wang2024sinsr} & {\color[HTML]{D83931} 26.30} & 0.7354 & 0.3212 & 0.2346 & 0.6204 & 6.31 & 60.41 & 0.5389 & 137.05 \\
                       & OSEDiff~\cite{wu2024one} & 25.15 & 0.7341 & 0.2921 & 0.2128 & 0.6693 & 5.65 & 69.09 & 0.6326 & 123.49 \\
                       & TVT~\cite{yi2025fine} & 25.81 & {\color[HTML]{D83931} 0.7597} & {\color[HTML]{245BDB} 0.2593} & 0.2061 & {\color[HTML]{245BDB} 0.6878} & 5.94 & 69.86 & 0.6228 & {\color[HTML]{245BDB} 110.07} \\
            {RealSR}  & TSD-SR~\cite{dong2025tsd} & 24.81 & 0.7172 & 0.2743 & 0.2104 & {\color[HTML]{D83931} 0.7160} & {\color[HTML]{D83931} 5.13} & {\color[HTML]{D83931} 71.19} & 0.6347 & 114.45 \\
                       & PiSA-SR~\cite{sun2025pixel} & 25.50 & 0.7417 & 0.2672 & {\color[HTML]{245BDB} 0.2044} & 0.6702 & 5.50 & {\color[HTML]{245BDB} 70.15} & {\color[HTML]{D83931} 0.6560} & 124.09 \\
                       & FSP-Diff & {\color[HTML]{245BDB} {\underline{26.09}}} & {\color[HTML]{245BDB} {\underline{0.7483}}} & {\color[HTML]{D83931} {\underline{0.2584}}} & {\color[HTML]{D83931} {\underline{0.1944}}} & 0.6687 & {\color[HTML]{245BDB}{\underline{5.26}}} & 68.60 & {\color[HTML]{245BDB} {\underline{0.6451}}} & {\color[HTML]{D83931} {\underline{108.46}}} \\ \hline
                       & SinSR~\cite{wang2024sinsr} & {\color[HTML]{245BDB} 24.43} & 0.6012 & 0.3262 & 0.2066 & 0.6499 & 6.02 & 62.80 & 0.5395 & 35.45 \\
                       & OSEDiff~\cite{wu2024one} & 23.72 & 0.6108 & 0.2941 & 0.1976 & 0.6683 & 4.71 & 67.97 & 0.6148 & 26.32 \\
                       & TVT~\cite{yi2025fine}& 24.23 & {\color[HTML]{D83931} 0.6292} & 0.2773 & {\color[HTML]{245BDB} 0.1860} & {\color[HTML]{245BDB}0.6987} & 5.60 & 68.67 & 0.6061 & {\color[HTML]{245BDB}24.79} \\
        {DIV2K-Val}      & TSD-SR~\cite{dong2025tsd} & 23.02 & 0.5808 & {\color[HTML]{D83931} 0.2673} & {\color[HTML]{D83931} 0.1821} & {\color[HTML]{D83931} 0.7416} & {\color[HTML]{D83931} 4.32} & {\color[HTML]{D83931} 71.69} & 0.6192 & 29.16 \\
                       & PiSA-SR~\cite{sun2025pixel} & 23.87 & 0.6058 & 0.2823 & 0.1934 & 0.6927 & {\color[HTML]{245BDB} 4.55} & {\color[HTML]{245BDB} 69.68} & {\color[HTML]{D83931} 0.6400} & 25.07 \\
                       & FSP-Diff & {\color[HTML]{D83931} {\underline{24.44}}} & {\color[HTML]{245BDB} {\underline{0.6278}}} & {\color[HTML]{245BDB} {\underline{0.2711}}} & {\underline{0.1873}} & 0.6583 & {\underline{4.60}} & 67.78 & {\color[HTML]{245BDB} {\underline{0.6283}}} & {\color[HTML]{D83931} {\underline{23.95}}} \\ \hline
\end{tabular}
}
\end{table*}
\noindent\textbf{Quantitative Comparisons.}
Table.~\ref{tab:quantity} presents a quantitative comparison of the proposed FSP-Diff method against state-of-the-art approaches on the Real-ISR task. Across both real-world and synthetic benchmarks, FSP-Diff consistently achieves superior performance, leading to the following observations: (1) Our model achieves clear improvements over OSEDiff on approximately 80\% of the evaluated metrics, while remaining competitive on the remaining 20\%. This demonstrates the effectiveness of introducing the proposed structured details via dual pathways.
(2) For reference-based metrics, FSP-Diff attains the highest frequency of best or second-best scores across both fidelity (PSNR, SSIM) and perceptual (LPIPS, DISTS) metrics, demonstrating its effectiveness in preserving fine structures while ensuring semantic coherence and perceptual realism. The FID metric records the lowest scores across all three benchmarks, suggesting strong distribution alignment between the generated images and their real-world counterparts.
(3) For no-reference metrics, our method achieves competitive or state-of-the-art performance on NIQE and MANIQA, indicating strong perceptual quality and naturalness. On CLIPIQA and MUSIQ, TVT and TSD-SR obtain higher scores, which may reflect their closer alignment with semantic-aware metrics derived from CLIP and MUSIQ. 
While these two methods report higher scores on CLIPIQA and MUSIQ, they employ additional training resources, such as complex multi-stage training or stronger pre-trained models. Compared to them, our approach adopts a simpler training configuration yet achieves competitive overall performance, reflecting a balanced trade-off between no-reference and reference-based metrics.

\noindent\textbf{Qualitative Comparisons.}
As shown in Fig.~\ref{fig:sota_comp}, we provide qualitative comparisons with the aforementioned state-of-the-art methods on the DIV2K-val benchmark.
Possibly arising from insufficient structured details, OSEDiff introduces higher stochasticity during generation and exhibits noticeable artifacts (e.g., incorrect blurring of the lion’s body in the second sample and skin tone inconsistencies in the third sample). TSD-SR shows limited ability to faithfully recover the original semantic content (e.g., missing lion details and deformed trains), likely due to an imbalance between distribution modeling and structured detail recovery, resulting in noticeable deviations from the input contents. TVT is a recent state-of-the-art Real-ISR model that, like our method, focuses on preserving fine structures and achieves this through its dedicated multi-stage training design. However, it may occasionally exhibit reduced perceptual naturalness(e.g., the most consistent pupil color in the third sample but lacking visual naturalness), indicating limited generation plausibility. In contrast, the proposed FSP-Diff, with its dual-pathway design, effectively preserves structured details while enabling flexible semantic refinement, leading to visually natural reconstructions with high detail fidelity.

\begin{figure}[tb]
  \centering
  \includegraphics[trim=0 0 0 0, clip, width=\linewidth]{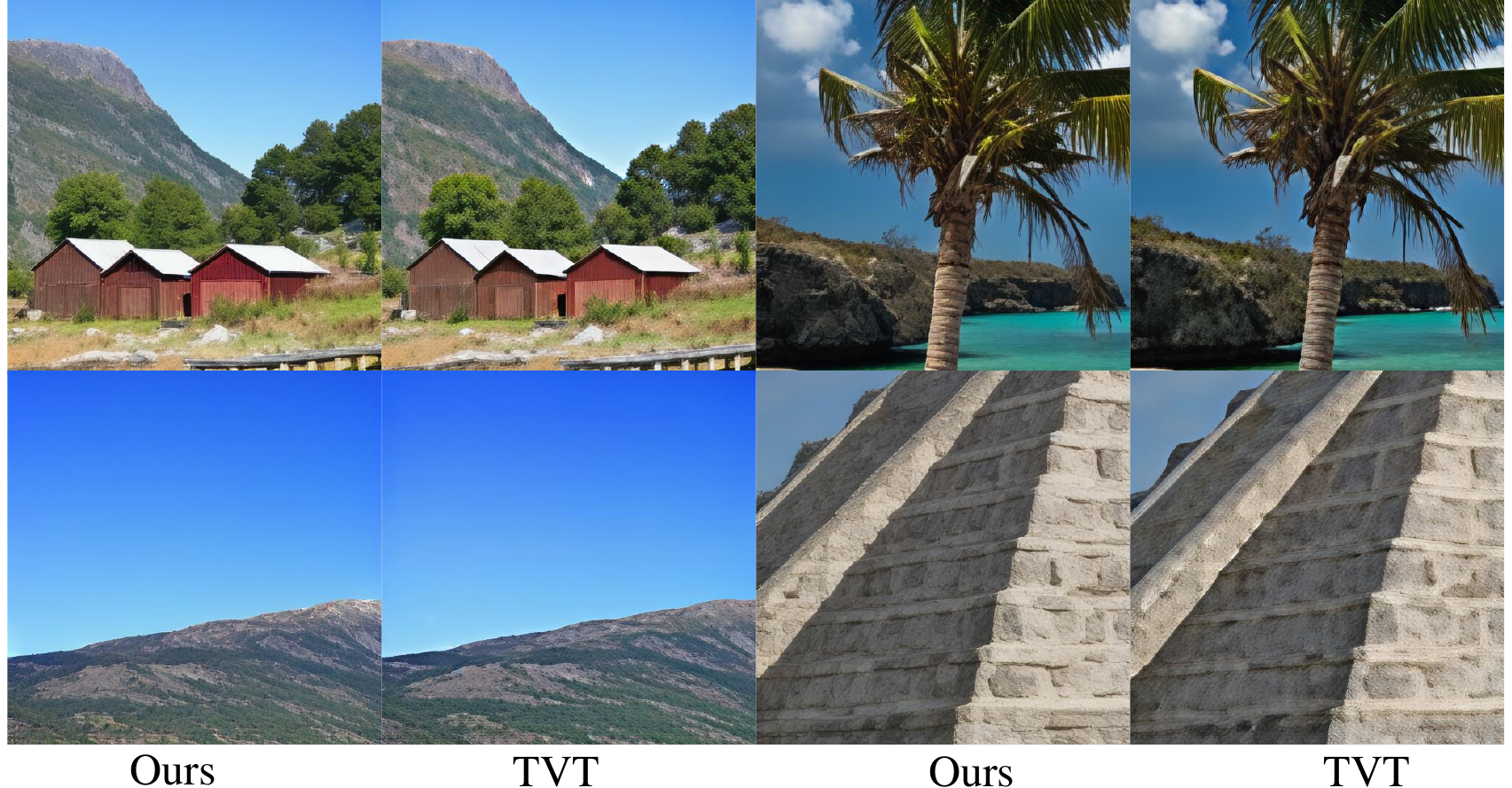}
  \caption{Comparison ofstructured detail-preserving Real-ISR methods: TVT versus our method. TVT tends to produce regular-pattern artifacts, such as unnatural polygonal curves on plants and buildings, whereas our method avoids or significantly reduces such artifacts. Please zoom in for a better view.
  }
  \label{fig:real_isr_tvt_comp}
\end{figure}

\begin{table*}[]
\centering
\caption{Ablation study on the selection of $E_f$ and the number of DATC blocks.}
\label{tab:ablation E_f}
\begin{tabular}{c|lclcllc}
\hline
\centering
Settings & SSIM↑ & DISTS↓ & CLIPIQA↑ & NIQE↓ & MUSIQ↑ & MANIQA↑ & FID↓ \\ \hline
\rowcolor[HTML]{FDE2E2}
ViT + DATC\_D4 & 0.7933 & \textbf{0.2014} & \textbf{0.6868} & 6.24 & \textbf{65.17} & \textbf{0.6092} & 127.19 \\
VAE + DATC\_D4 & 0.7933 & 0.2048 & 0.6725 & \textbf{6.21} & 64.95 & 0.6053 & \textbf{125.34} \\
\multicolumn{1}{l|}{ViT + DATC\_D1} & \textbf{0.7998} & \multicolumn{1}{l}{0.2017} & 0.6672 & 6.47 & 63.74 & 0.6000 & \multicolumn{1}{l}{127.13} \\
VAE + DATC\_D1 & 0.7933 & \multicolumn{1}{l}{0.2043} & 0.6862 & 6.41 & 64.76 & 0.6048 & \multicolumn{1}{l}{125.79} \\ \hline
\end{tabular}
\end{table*}
To further compare our method with TVT, we provide additional qualitative comparisons. While TVT preserves structured details through its specially designed VAE-D4~\cite{yi2025fine} features, it produces noticeable rigid artifacts in certain regions. For example, as illustrated in Fig.~\ref{fig:real_isr_tvt_comp}, mountains, grass, and city walls are filled with regular polygonal artifacts, which even propagate to parts of the human figures shown in the first example of the second row. These artifacts may stem from the direct residual injection ofstructured detail features before the decoder, which lacks effective refinement and denoising, leading to unnatural details. In contrast, by appropriately compressing and leveraging structured details, our method effectively avoids such noisy patterns, producing more natural and faithful reconstructions.

\subsection{Ablation Study}

\noindent\textbf{Selection of Image Extractor $E_f$.}
We explore two selections for the feature extractor $E_f$: (1) a shallow convolutional backbone and (2) the early layers of a ViT-based encoder. The former adopts the same architecture and pre-trained weights as $E_{\theta}$ and uses the feature from the third downsampling stage, which balances spatial resolution and semantic abstraction. The latter adopts the first block of the image encoder in ``$vit\_b$''\cite{dosovitskiy2020image}, corresponding to the method described in Section\ref{sec:tcp}. In Table.~\ref{tab:ablation E_f}, “VAE+” and “ViT+” correspond to the two selections, respectively. “DATC\_DN” denotes stacking $N$ blocks in the DATC module. We observe that “ViT+DATC\_D4” achieves the best performance on most metrics, indicating that shallow features extracted from a pre-trained ViT are more effective for structured details extraction than those from convolutional neural networks.

\noindent\textbf{Ablation on ViT depth.}
Since early-layer ViT features best preserve the fine-grained structural details degraded by VAE compression, we further vary the ViT depth of $E_f$ and report the results on DRealSR in Table.~\ref{tab:vit}. Using the first block achieves the best performance across all metrics, validating our choice of the frozen ViT's first block as the structural prior.

\noindent\textbf{Effectiveness of DATC and SDCA.}
To verify the effectiveness of the key components in the Detail-Conditioned Pathway, we perform ablation studies on the OSEDiff baseline by progressively adding SDCA and DATC. The corresponding results, evaluated on the DRealSR test set, are highlighted with blue background in Table.~\ref{tab:ablation1}. We note that although the best-performing model in Table.~\ref{tab:ablation E_f} adopts the “ViT+” selection for structured details extracting, the following ablation studies are conducted with the “VAE+” image extractor setting. This choice does not affect the conclusions of the component-wise ablation analysis.

We evaluate all variants using PSNR, DISTS, NIQE, and MANIQA, covering both fidelity and perceptual quality, as well as reference-based and no-reference metrics. In addition, FID is employed to assess distribution similarity. The setting "Add SDCA" denotes introducing only the SDCA module into the OSEDiff baseline, where the extracted structured details is directly processed by a fixed 4× downsampling, then fed as input to SDCA. Based on this, "Add DATC" replaces the fixed 4× downsampling with our proposed DATC module, resulting in the complete Detail-Conditioned Pathway design. From the results, we draw the following conclusions: (1)According to the "Add SDCA" experiment, introducing the structured details via SDCA into the SD-based model improves most of the evaluated metrics, showing that this structured prior effectively enhances both fidelity and perceptual quality. (2)According to the "Add DATC" experiment, DATC more effectively compresses and refines structured details features compared with fixed 4× downsampling, preserving fine structures for HQ image generation while reducing tokens.

 \begin{table}[htbp]
      \centering
      \small
      \caption{Ablation on ViT depth on DrealSR Benchmark.}
      \label{tab:vit}
        \resizebox{\linewidth}{!}{
        \begin{tabular}{lccccc}
            \toprule
            ViT Depth & LPIPS$\downarrow$ & DISTS$\downarrow$ & NIQE$\downarrow$ & MUSIQ$\uparrow$ & MANIQA$\uparrow$ \\
            \midrule
            1 & \textbf{0.2741} & \textbf{0.2014} & \textbf{6.24} & \textbf{65.17} & \textbf{0.6092} \\
            2 & 0.2775 & 0.2037 & 6.49 & 63.05 & 0.5930 \\
            3 & 0.2764 & 0.2096 & 6.72 & 64.74 & 0.5937 \\
            \bottomrule
      \end{tabular}}
  \end{table}

\begin{table}[]
\centering
\caption{Ablation Study on Key Components of FSP-Diff. "Baseline" and "Full" correspond to results
from the baseline model~\cite{wu2024one} without our proposed modules and
FSP-Diff with all components enabled, respectively.}
\label{tab:ablation1}
\begin{tabular}{l|lllll}
\hline
Methods & PSNR↑ & DISTS↓ & NIQE↓ & MANIQA↑ & FID↓ \\ \hline
Baseline & 27.92 & 0.2165 & 6.49 & 0.5899 & 135.30 \\
\rowcolor[HTML]{DAE8FC} 
add SDCA & 28.60 & 0.2109 & 6.82 & 0.5998 & 133.49 \\
\rowcolor[HTML]{DAE8FC}
add DATC & {\color[HTML]{333333} 28.90} & 0.2073 & 6.58 & 0.5975 & 128.66 \\
\rowcolor[HTML]{DAE8FC} 
add SDM & \textbf{29.06} & \textbf{0.2048} & 6.45 & 0.6000 & \textbf{124.55} \\
\rowcolor[HTML]{FDE2E2}
Full & 28.93 & \textbf{0.2048} & \textbf{6.21} & \textbf{0.6053} & 125.34 \\ \hline
\end{tabular}
\end{table}

\begin{table}[htbp]
    \centering
    
    \caption{Ablation of $L_q$ for DATC.}
    \label{tab:ablation lq}
    \begin{tabular}[t]{l|lllll}
    \hline
    $L_q$ & PSNR↑ & DISTS↓ & MUSIQ↑ & MANIQA↑ & FID↓ \\ \hline
    300 & 28.88 & \textbf{0.2028} & 63.84 & 0.5885 & 125.45 \\
    \rowcolor[HTML]{FDE2E2} 
    400 & \textbf{28.93} & 0.2048 & \textbf{64.95} & \textbf{0.6053} & \textbf{125.34} \\ 
    500 & 28.91 & 0.2031 & 64.43 & 0.5999 & 127.64 \\ \hline
    \end{tabular}
\end{table}
\begin{table}[htbp]
   
    \centering
    \caption{Impact of shift parameters $s$ on DIV2K Benchmark.}
    \label{tab:ablation s}
     \begin{tabular}{lcccccc}
        \toprule
        $s$ & LPIPS$\downarrow$ & DISTS$\downarrow$ & NIQE$\downarrow$ & MUSIQ$\uparrow$ & MANIQA$\uparrow$ & FID$\downarrow$ \\
        \midrule
        0.9 & \textbf{0.2718} & \textbf{0.1896} & \textbf{4.60} & \textbf{67.69} & \textbf{0.6267} & \textbf{24.15} \\
        1.0 & 0.2782 & 0.1920 & 4.67 & 67.58 & 0.6248 & 24.53 \\
        1.2 & 0.2740 & 0.1899 & 4.74 & 67.15 & 0.6252 & 24.48 \\
        \bottomrule
    \end{tabular}
    
\end{table}

\noindent\textbf{$L_q$ setting for DATC module.}
As described in Section~\ref{sec:tcp}, we analyze different settings of the learnable query length
$L_q$ in the DATC module. A larger $L_q$ increases computational cost, as the learnable queries participate in the attention operations of the SDCA module. As shown in Table.~\ref{tab:ablation lq}, setting $L_q$=400 provides a favorable trade-off between performance and efficiency. A smaller $L_q$ compromises the preservation of structured details, while a larger $L_q$ yields negligible performance gains and increases computational cost.


\noindent\textbf{Effectiveness of SDM.}
As indicated by the third blue rows in Table.~\ref{tab:ablation1}, "Add SDM" denotes the complete dual-pathway architecture, formed by integrating the SDM module into the existing “Add DATC” configuration. The results show consistent improvements in both reference-based and no-reference metrics, which is non-trivial as these metrics typically require careful trade-offs. This demonstrates that integrating structured details can enhance text-driven semantic guidance, enabling the refined semantics to more effectively exploit generation priors in T2I models.

\noindent\textbf{Impact of the Shift Hyperparameter.}
We explore a shift operation on the noise schedulers of the two regularizers, $\varepsilon_{\phi}$ and $\varepsilon_{\phi'}$. The shift hyperparameter s is constrained to a narrow range around 1, as larger deviations cause training instability, likely due to excessive distillation at extreme noise levels diverting the student model from the intended distribution. Table.~\ref{tab:ablation s} compares several discrete $s$ values on the DRealSR test set. Among the tested values, $s$=0.9 yields the best overall scores, and we adopt adopt this value as a stable default. In Table.~\ref{tab:ablation1}, the only difference between experimemts "Add SDM" and "Full" lies in $s$: "Add SDM" uses $s$=1.0 (equivalent to no shift), while "Full" uses $s$=0.9. 
\begin{figure}[tb]
  \centering
  \includegraphics[trim=0 43 0 0, clip, width=\linewidth]{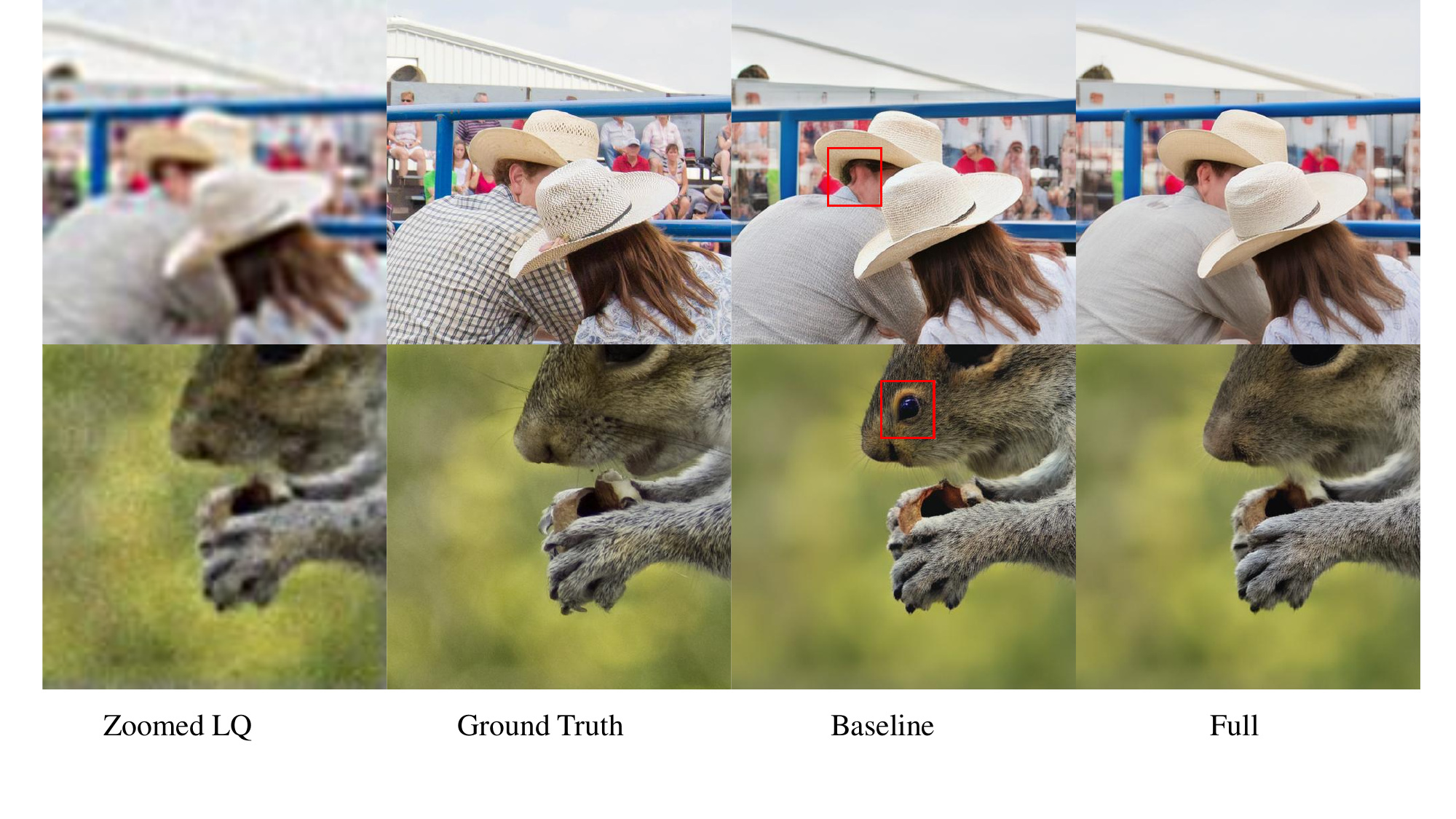}
  \caption{Qualitative ablation comparison. Red boxes highlight regions with semantic deviations in the generated results. "Baseline" and "Full" correspond to results from the OSEDiff baseline without our proposed modules and FSP-Diff with all components enabled, respectively.
  }
  \label{fig:real-sr ablation}
\end{figure}

\noindent\textbf{Qualitative Ablation Results.}
To provide a more intuitive comparison between our model and the baseline, we present qualitative results in Fig.~\ref{fig:real-sr ablation}, corresponding to the quantitative improvements reported in Table.~\ref{tab:ablation1} , where “Full” denotes our best configuration with SDCA, DATC, and SDM, and the shift parameter set to 0.9. As shown in the examples, the baseline model exhibits semantic misinterpretation of the LQ input, which can lead to incorrect fine structures in the generated HQ images, as highlighted by the red boxes. In contrast, our full model preserves fine-grained information, mitigating such semantic deviations and producing visually consistent and plausible results.

\noindent\textbf{Complexity Comparisons.}
In Table.~\ref{tab:param, flops and time}, we compare the computational complexity of FSP-Diff with competing diffusion-based Real-ISR methods, including the total number of parameters, FLOPs and inference time. All inference times are measured on an A100 using $512 \times 512$ input images. Although our method introduces a dual-pathway architecture based on the OSEDiff model, it incurs only a modest increase in computational cost while maintaining significantly lower FLOPs and inference time than the multi-step diffusion models, such as StableSR~\cite{wang2024exploiting} and SeeSR~\cite{wu2024seesr}, as shown in Table.~\ref{tab:param, flops and time}. Moreover, despite having more parameters than the current state-of-the-art one-step model TVT~\cite{yi2025fine}, our method achieves lower inference latency.
\begin{table}[]
\caption{Comparison of parameter count (Para.), FLOPs and inference time(Time).}
\label{tab:param, flops and time}
\resizebox{\linewidth}{!}{
\begin{tabular}{c|lclc|lll}
\hline
  & StableSR & DiffBIR & SeeSR & PASD & OSEDiff & TVT & FSP-Diff \\ \hline
Step & 200 & 50 & 50 & 20 & 1 & 1 & 1 \\
Para. (B) & 1.56 & 1.68 & 2.51 & 2.31 & 1.77 & 1.72 & 2.06  \\
\multicolumn{1}{l|}{FLOPs (T)} & 79.94 & \multicolumn{1}{l}{24.31} & 65.86 & \multicolumn{1}{l|}{29.13} & 2.16 & 1.97 & 2.44  \\ 
Time (s) & 12.42 & 7.96 & 5.82 & 4.84 & 0.12 & 0.24 & 0.15  \\
\hline
\end{tabular}
}
\end{table}

\section{Conclusion}
In this paper, we proposed FSP-Diff, a novel one-step diffusion framework for real-world image super-resolution that performs fine-grained generation in an end-to-end manner. Central to our approach is the extraction of fine-grained structural priors from low-quality inputs to preserve structured details. We incorporated these priors into a dual-pathway architecture: (1) a Detail-Conditioned Pathway, which compresses and transforms structured details for precise fine-grained detail integration, and (2) a Detail-Modulated Semantic Pathway, which regulates semantic guidance based on structured details to mitigate semantic inconsistencies. Extensive experiments demonstrate that FSP-Diff achieves competitive or superior performance compared to state-of-the-art methods in both quantitative metrics and qualitative visualizations. These results validate the effectiveness of our framework in preserving fine structures and enhancing the robustness of semantic guidance in Real-ISR.




\bibliographystyle{ACM-Reference-Format}
\bibliography{sigconf}

@String{Computing = "Computing" }

@String{Computer = "{IEEE} Computer" }

@String{Springer = "Springer-Verlag" }

@inproceedings{zhang2021designing,
  title={Designing a practical degradation model for deep blind image super-resolution},
  author={Zhang, Kai and Liang, Jingyun and Van Gool, Luc and Timofte, Radu},
  booktitle={Proceedings of the IEEE/CVF international conference on computer vision},
  pages={4791--4800},
  year={2021}
}

@inproceedings{wang2021real,
  title={Real-esrgan: Training real-world blind super-resolution with pure synthetic data},
  author={Wang, Xintao and Xie, Liangbin and Dong, Chao and Shan, Ying},
  booktitle={Proceedings of the IEEE/CVF international conference on computer vision},
  pages={1905--1914},
  year={2021}
}

@inproceedings{chen2021pre,
  title={Pre-trained image processing transformer},
  author={Chen, Hanting and Wang, Yunhe and Guo, Tianyu and Xu, Chang and Deng, Yiping and Liu, Zhenhua and Ma, Siwei and Xu, Chunjing and Xu, Chao and Gao, Wen},
  booktitle={Proceedings of the IEEE/CVF conference on computer vision and pattern recognition},
  pages={12299--12310},
  year={2021}
}

@inproceedings{zhang2022efficient,
  title={Efficient long-range attention network for image super-resolution},
  author={Zhang, Xindong and Zeng, Hui and Guo, Shi and Zhang, Lei},
  booktitle={European conference on computer vision},
  pages={649--667},
  year={2022},
  organization={Springer}
}

@inproceedings{dai2019second,
  title={Second-order attention network for single image super-resolution},
  author={Dai, Tao and Cai, Jianrui and Zhang, Yongbing and Xia, Shu-Tao and Zhang, Lei},
  booktitle={Proceedings of the IEEE/CVF conference on computer vision and pattern recognition},
  pages={11065--11074},
  year={2019}
}

@inproceedings{liang2021swinir,
  title={Swinir: Image restoration using swin transformer},
  author={Liang, Jingyun and Cao, Jiezhang and Sun, Guolei and Zhang, Kai and Van Gool, Luc and Timofte, Radu},
  booktitle={Proceedings of the IEEE/CVF international conference on computer vision},
  pages={1833--1844},
  year={2021}
}

@inproceedings{zhang2018image,
  title={Image super-resolution using very deep residual channel attention networks},
  author={Zhang, Yulun and Li, Kunpeng and Li, Kai and Wang, Lichen and Zhong, Bineng and Fu, Yun},
  booktitle={Proceedings of the European conference on computer vision (ECCV)},
  pages={286--301},
  year={2018}
}

@inproceedings{wei2020component,
  title={Component divide-and-conquer for real-world image super-resolution},
  author={Wei, Pengxu and Xie, Ziwei and Lu, Hannan and Zhan, Zongyuan and Ye, Qixiang and Zuo, Wangmeng and Lin, Liang},
  booktitle={European conference on computer vision},
  pages={101--117},
  year={2020},
  organization={Springer}
}

@inproceedings{cai2019toward,
  title={Toward real-world single image super-resolution: A new benchmark and a new model},
  author={Cai, Jianrui and Zeng, Hui and Yong, Hongwei and Cao, Zisheng and Zhang, Lei},
  booktitle={Proceedings of the IEEE/CVF international conference on computer vision},
  pages={3086--3095},
  year={2019}
}

@article{saharia2022photorealistic,
  title={Photorealistic text-to-image diffusion models with deep language understanding},
  author={Saharia, Chitwan and Chan, William and Saxena, Saurabh and Li, Lala and Whang, Jay and Denton, Emily L and Ghasemipour, Kamyar and Gontijo Lopes, Raphael and Karagol Ayan, Burcu and Salimans, Tim and others},
  journal={Advances in neural information processing systems},
  volume={35},
  pages={36479--36494},
  year={2022}
}

@inproceedings{rombach2022high,
  title={High-resolution image synthesis with latent diffusion models},
  author={Rombach, Robin and Blattmann, Andreas and Lorenz, Dominik and Esser, Patrick and Ommer, Bj{\"o}rn},
  booktitle={Proceedings of the IEEE/CVF conference on computer vision and pattern recognition},
  pages={10684--10695},
  year={2022}
}

@inproceedings{esser2024scaling,
  title={Scaling rectified flow transformers for high-resolution image synthesis},
  author={Esser, Patrick and Kulal, Sumith and Blattmann, Andreas and Entezari, Rahim and M{\"u}ller, Jonas and Saini, Harry and Levi, Yam and Lorenz, Dominik and Sauer, Axel and Boesel, Frederic and others},
  booktitle={Forty-first international conference on machine learning},
  year={2024}
}

@inproceedings{wu2024seesr,
  title={Seesr: Towards semantics-aware real-world image super-resolution},
  author={Wu, Rongyuan and Yang, Tao and Sun, Lingchen and Zhang, Zhengqiang and Li, Shuai and Zhang, Lei},
  booktitle={Proceedings of the IEEE/CVF conference on computer vision and pattern recognition},
  pages={25456--25467},
  year={2024}
}

@inproceedings{yang2024pixel,
  title={Pixel-aware stable diffusion for realistic image super-resolution and personalized stylization},
  author={Yang, Tao and Wu, Rongyuan and Ren, Peiran and Xie, Xuansong and Zhang, Lei},
  booktitle={European conference on computer vision},
  pages={74--91},
  year={2024},
  organization={Springer}
}

@inproceedings{agustsson2017ntire,
  title={Ntire 2017 challenge on single image super-resolution: Dataset and study},
  author={Agustsson, Eirikur and Timofte, Radu},
  booktitle={Proceedings of the IEEE conference on computer vision and pattern recognition workshops},
  pages={126--135},
  year={2017}
}

@inproceedings{chen2022real,
  title={Real-world blind super-resolution via feature matching with implicit high-resolution priors},
  author={Chen, Chaofeng and Shi, Xinyu and Qin, Yipeng and Li, Xiaoming and Han, Xiaoguang and Yang, Tao and Guo, Shihui},
  booktitle={Proceedings of the 30th ACM International Conference on Multimedia},
  pages={1329--1338},
  year={2022}
}

@inproceedings{liang2022details,
  title={Details or artifacts: A locally discriminative learning approach to realistic image super-resolution},
  author={Liang, Jie and Zeng, Hui and Zhang, Lei},
  booktitle={Proceedings of the IEEE/CVF conference on computer vision and pattern recognition},
  pages={5657--5666},
  year={2022}
}

@inproceedings{liang2022efficient,
  title={Efficient and degradation-adaptive network for real-world image super-resolution},
  author={Liang, Jie and Zeng, Hui and Zhang, Lei},
  booktitle={European Conference on Computer Vision},
  pages={574--591},
  year={2022},
  organization={Springer}
}

@article{xie2023desra,
  title={Desra: detect and delete the artifacts of gan-based real-world super-resolution models},
  author={Xie, Liangbin and Wang, Xintao and Chen, Xiangyu and Li, Gen and Shan, Ying and Zhou, Jiantao and Dong, Chao},
  journal={arXiv preprint arXiv:2307.02457},
  year={2023}
}

@article{ho2020denoising,
  title={Denoising diffusion probabilistic models},
  author={Ho, Jonathan and Jain, Ajay and Abbeel, Pieter},
  journal={Advances in neural information processing systems},
  volume={33},
  pages={6840--6851},
  year={2020}
}

@article{song2020denoising,
  title={Denoising diffusion implicit models},
  author={Song, Jiaming and Meng, Chenlin and Ermon, Stefano},
  journal={arXiv preprint arXiv:2010.02502},
  year={2020}
}

@article{dhariwal2021diffusion,
  title={Diffusion models beat gans on image synthesis},
  author={Dhariwal, Prafulla and Nichol, Alexander},
  journal={Advances in neural information processing systems},
  volume={34},
  pages={8780--8794},
  year={2021}
}

@article{wang2024exploiting,
  title={Exploiting diffusion prior for real-world image super-resolution},
  author={Wang, Jianyi and Yue, Zongsheng and Zhou, Shangchen and Chan, Kelvin CK and Loy, Chen Change},
  journal={International Journal of Computer Vision},
  volume={132},
  number={12},
  pages={5929--5949},
  year={2024},
  publisher={Springer}
}

@article{yue2023resshift,
  title={Resshift: Efficient diffusion model for image super-resolution by residual shifting},
  author={Yue, Zongsheng and Wang, Jianyi and Loy, Chen Change},
  journal={Advances in Neural Information Processing Systems},
  volume={36},
  pages={13294--13307},
  year={2023}
}

@article{kawar2022denoising,
  title={Denoising diffusion restoration models},
  author={Kawar, Bahjat and Elad, Michael and Ermon, Stefano and Song, Jiaming},
  journal={Advances in neural information processing systems},
  volume={35},
  pages={23593--23606},
  year={2022}
}

@inproceedings{lin2024diffbir,
  title={Diffbir: Toward blind image restoration with generative diffusion prior},
  author={Lin, Xinqi and He, Jingwen and Chen, Ziyan and Lyu, Zhaoyang and Dai, Bo and Yu, Fanghua and Qiao, Yu and Ouyang, Wanli and Dong, Chao},
  booktitle={European conference on computer vision},
  pages={430--448},
  year={2024},
  organization={Springer}
}

@inproceedings{qu2024xpsr,
  title={Xpsr: Cross-modal priors for diffusion-based image super-resolution},
  author={Qu, Yunpeng and Yuan, Kun and Zhao, Kai and Xie, Qizhi and Hao, Jinhua and Sun, Ming and Zhou, Chao},
  booktitle={European Conference on Computer Vision},
  pages={285--303},
  year={2024},
  organization={Springer}
}

@inproceedings{sun2025pixel,
  title={Pixel-level and semantic-level adjustable super-resolution: A dual-lora approach},
  author={Sun, Lingchen and Wu, Rongyuan and Ma, Zhiyuan and Liu, Shuaizheng and Yi, Qiaosi and Zhang, Lei},
  booktitle={Proceedings of the Computer Vision and Pattern Recognition Conference},
  pages={2333--2343},
  year={2025}
}

@article{xie2024addsr,
  title={Addsr: Accelerating diffusion-based blind super-resolution with adversarial diffusion distillation},
  author={Xie, Rui and Zhao, Chen and Zhang, Kai and Zhang, Zhenyu and Zhou, Jun and Yang, Jian and Tai, Ying},
  journal={arXiv preprint arXiv:2404.01717},
  year={2024}
}

@article{wu2024one,
  title={One-step effective diffusion network for real-world image super-resolution},
  author={Wu, Rongyuan and Sun, Lingchen and Ma, Zhiyuan and Zhang, Lei},
  journal={Advances in Neural Information Processing Systems},
  volume={37},
  pages={92529--92553},
  year={2024}
}

@inproceedings{chen2025faithdiff,
  title={Faithdiff: Unleashing diffusion priors for faithful image super-resolution},
  author={Chen, Junyang and Pan, Jinshan and Dong, Jiangxin},
  booktitle={Proceedings of the Computer Vision and Pattern Recognition Conference},
  pages={28188--28197},
  year={2025}
}

@inproceedings{dong2025tsd,
  title={Tsd-sr: One-step diffusion with target score distillation for real-world image super-resolution},
  author={Dong, Linwei and Fan, Qingnan and Guo, Yihong and Wang, Zhonghao and Zhang, Qi and Chen, Jinwei and Luo, Yawei and Zou, Changqing},
  booktitle={Proceedings of the Computer Vision and Pattern Recognition Conference},
  pages={23174--23184},
  year={2025}
}

@inproceedings{yi2025fine,
  title={Fine-structure preserved real-world image super-resolution via transfer vae training},
  author={Yi, Qiaosi and Li, Shuai and Wu, Rongyuan and Sun, Lingchen and Wu, Yuhui and Zhang, Lei},
  booktitle={Proceedings of the IEEE/CVF international conference on computer vision},
  pages={12415--12426},
  year={2025}
}

@article{zhang2024degradation,
  title={Degradation-guided one-step image super-resolution with diffusion priors},
  author={Zhang, Aiping and Yue, Zongsheng and Pei, Renjing and Ren, Wenqi and Cao, Xiaochun},
  journal={arXiv preprint arXiv:2409.17058},
  year={2024}
}

@inproceedings{yu2024scaling,
  title={Scaling up to excellence: Practicing model scaling for photo-realistic image restoration in the wild},
  author={Yu, Fanghua and Gu, Jinjin and Li, Zheyuan and Hu, Jinfan and Kong, Xiangtao and Wang, Xintao and He, Jingwen and Qiao, Yu and Dong, Chao},
  booktitle={Proceedings of the IEEE/CVF conference on computer vision and pattern recognition},
  pages={25669--25680},
  year={2024}
}

@article{podell2023sdxl,
  title={Sdxl: Improving latent diffusion models for high-resolution image synthesis},
  author={Podell, Dustin and English, Zion and Lacey, Kyle and Blattmann, Andreas and Dockhorn, Tim and M{\"u}ller, Jonas and Penna, Joe and Rombach, Robin},
  journal={arXiv preprint arXiv:2307.01952},
  year={2023}
}

@article{chen2023pixart,
  title={$\text{Pixart-}\alpha$: Fast training of diffusion transformer for photorealistic text-to-image synthesis},
  author={Chen, Junsong and Yu, Jincheng and Ge, Chongjian and Yao, Lewei and Xie, Enze and Wu, Yue and Wang, Zhongdao and Kwok, James and Luo, Ping and Lu, Huchuan and others},
  journal={arXiv preprint arXiv:2310.00426},
  year={2023}
}

@article{wang2023prolificdreamer,
  title={Prolificdreamer: High-fidelity and diverse text-to-3d generation with variational score distillation},
  author={Wang, Zhengyi and Lu, Cheng and Wang, Yikai and Bao, Fan and Li, Chongxuan and Su, Hang and Zhu, Jun},
  journal={Advances in neural information processing systems},
  volume={36},
  pages={8406--8441},
  year={2023}
}

@inproceedings{yin2024one,
  title={One-step diffusion with distribution matching distillation},
  author={Yin, Tianwei and Gharbi, Micha{\"e}l and Zhang, Richard and Shechtman, Eli and Durand, Fredo and Freeman, William T and Park, Taesung},
  booktitle={Proceedings of the IEEE/CVF conference on computer vision and pattern recognition},
  pages={6613--6623},
  year={2024}
}

@inproceedings{dao2024swiftbrush,
  title={Swiftbrush v2: Make your one-step diffusion model better than its teacher},
  author={Dao, Trung and Nguyen, Thuan Hoang and Le, Thanh and Vu, Duc and Nguyen, Khoi and Pham, Cuong and Tran, Anh},
  booktitle={European Conference on Computer Vision},
  pages={176--192},
  year={2024},
  organization={Springer}
}

@inproceedings{wang2024sinsr,
  title={Sinsr: diffusion-based image super-resolution in a single step},
  author={Wang, Yufei and Yang, Wenhan and Chen, Xinyuan and Wang, Yaohui and Guo, Lanqing and Chau, Lap-Pui and Liu, Ziwei and Qiao, Yu and Kot, Alex C and Wen, Bihan},
  booktitle={Proceedings of the IEEE/CVF conference on computer vision and pattern recognition},
  pages={25796--25805},
  year={2024}
}

@article{liu2023visual,
  title={Visual instruction tuning},
  author={Liu, Haotian and Li, Chunyuan and Wu, Qingyang and Lee, Yong Jae},
  journal={Advances in neural information processing systems},
  volume={36},
  pages={34892--34916},
  year={2023}
}

@article{kuznetsova2020open,
  title={The open images dataset v4: Unified image classification, object detection, and visual relationship detection at scale},
  author={Kuznetsova, Alina and Rom, Hassan and Alldrin, Neil and Uijlings, Jasper and Krasin, Ivan and Pont-Tuset, Jordi and Kamali, Shahab and Popov, Stefan and Malloci, Matteo and Kolesnikov, Alexander and others},
  journal={International journal of computer vision},
  volume={128},
  number={7},
  pages={1956--1981},
  year={2020},
  publisher={Springer}
}

@inproceedings{zhang2023adding,
  title={Adding conditional control to text-to-image diffusion models},
  author={Zhang, Lvmin and Rao, Anyi and Agrawala, Maneesh},
  booktitle={Proceedings of the IEEE/CVF international conference on computer vision},
  pages={3836--3847},
  year={2023}
}

@inproceedings{lim2017enhanced,
  title={Enhanced deep residual networks for single image super-resolution},
  author={Lim, Bee and Son, Sanghyun and Kim, Heewon and Nah, Seungjun and Mu Lee, Kyoung},
  booktitle={Proceedings of the IEEE conference on computer vision and pattern recognition workshops},
  pages={136--144},
  year={2017}
}

@article{goodfellow2014generative,
  title={Generative adversarial nets},
  author={Goodfellow, Ian J and Pouget-Abadie, Jean and Mirza, Mehdi and Xu, Bing and Warde-Farley, David and Ozair, Sherjil and Courville, Aaron and Bengio, Yoshua},
  journal={Advances in neural information processing systems},
  volume={27},
  year={2014}
}

@article{song2020score,
  title={Score-based generative modeling through stochastic differential equations},
  author={Song, Yang and Sohl-Dickstein, Jascha and Kingma, Diederik P and Kumar, Abhishek and Ermon, Stefano and Poole, Ben},
  journal={arXiv preprint arXiv:2011.13456},
  year={2020}
}

@article{hu2022lora,
  title={Lora: Low-rank adaptation of large language models.},
  author={Hu, Edward J and Shen, Yelong and Wallis, Phillip and Allen-Zhu, Zeyuan and Li, Yuanzhi and Wang, Shean and Wang, Lu and Chen, Weizhu and others},
  journal={ICLR},
  volume={1},
  number={2},
  pages={3},
  year={2022}
}

@article{luo2023latent,
  title={Latent consistency models: Synthesizing high-resolution images with few-step inference},
  author={Luo, Simian and Tan, Yiqin and Huang, Longbo and Li, Jian and Zhao, Hang},
  journal={arXiv preprint arXiv:2310.04378},
  year={2023}
}

@inproceedings{wang2025osdface,
  title={Osdface: One-step diffusion model for face restoration},
  author={Wang, Jingkai and Gong, Jue and Zhang, Lin and Chen, Zheng and Liu, Xing and Gu, Hong and Liu, Yutong and Zhang, Yulun and Yang, Xiaokang},
  booktitle={Proceedings of the Computer Vision and Pattern Recognition Conference},
  pages={12626--12636},
  year={2025}
}

@inproceedings{zhang2018unreasonable,
  title={The unreasonable effectiveness of deep features as a perceptual metric},
  author={Zhang, Richard and Isola, Phillip and Efros, Alexei A and Shechtman, Eli and Wang, Oliver},
  booktitle={Proceedings of the IEEE conference on computer vision and pattern recognition},
  pages={586--595},
  year={2018}
}

@article{vaswani2017attention,
  title={Attention is all you need},
  author={Vaswani, Ashish and Shazeer, Noam and Parmar, Niki and Uszkoreit, Jakob and Jones, Llion and Gomez, Aidan N and Kaiser, {\L}ukasz and Polosukhin, Illia},
  journal={Advances in neural information processing systems},
  volume={30},
  year={2017}
}

@inproceedings{li2023lsdir,
  title={Lsdir: A large scale dataset for image restoration},
  author={Li, Yawei and Zhang, Kai and Liang, Jingyun and Cao, Jiezhang and Liu, Ce and Gong, Rui and Zhang, Yulun and Tang, Hao and Liu, Yun and Demandolx, Denis and others},
  booktitle={Proceedings of the IEEE/CVF Conference on Computer Vision and Pattern Recognition},
  pages={1775--1787},
  year={2023}
}

@inproceedings{karras2019style,
  title={A style-based generator architecture for generative adversarial networks},
  author={Karras, Tero and Laine, Samuli and Aila, Timo},
  booktitle={Proceedings of the IEEE/CVF conference on computer vision and pattern recognition},
  pages={4401--4410},
  year={2019}
}

@inproceedings{ma2023benchmark,
  title={A benchmark for chinese-english scene text image super-resolution},
  author={Ma, Jianqi and Liang, Zhetong and Xiang, Wangmeng and Yang, Xi and Zhang, Lei},
  booktitle={Proceedings of the IEEE/CVF International Conference on Computer Vision},
  pages={19452--19461},
  year={2023}
}

@inproceedings{timofte2017ntire,
  title={Ntire 2017 challenge on single image super-resolution: Methods and results},
  author={Timofte, Radu and Agustsson, Eirikur and Van Gool, Luc and Yang, Ming-Hsuan and Zhang, Lei},
  booktitle={Proceedings of the IEEE conference on computer vision and pattern recognition workshops},
  pages={114--125},
  year={2017}
}

@article{wang2004image,
  title={Image quality assessment: from error visibility to structural similarity},
  author={Wang, Zhou and Bovik, Alan C and Sheikh, Hamid R and Simoncelli, Eero P},
  journal={IEEE transactions on image processing},
  volume={13},
  number={4},
  pages={600--612},
  year={2004},
  publisher={IEEE}
}

@article{ding2020image,
  title={Image quality assessment: Unifying structure and texture similarity},
  author={Ding, Keyan and Ma, Kede and Wang, Shiqi and Simoncelli, Eero P},
  journal={IEEE transactions on pattern analysis and machine intelligence},
  volume={44},
  number={5},
  pages={2567--2581},
  year={2020},
  publisher={IEEE}
}

@article{heusel2017gans,
  title={Gans trained by a two time-scale update rule converge to a local nash equilibrium},
  author={Heusel, Martin and Ramsauer, Hubert and Unterthiner, Thomas and Nessler, Bernhard and Hochreiter, Sepp},
  journal={Advances in neural information processing systems},
  volume={30},
  year={2017}
}

@article{zhang2015feature,
  title={A feature-enriched completely blind image quality evaluator},
  author={Zhang, Lin and Zhang, Lei and Bovik, Alan C},
  journal={IEEE Transactions on Image Processing},
  volume={24},
  number={8},
  pages={2579--2591},
  year={2015},
  publisher={IEEE}
}

@inproceedings{yang2022maniqa,
  title={Maniqa: Multi-dimension attention network for no-reference image quality assessment},
  author={Yang, Sidi and Wu, Tianhe and Shi, Shuwei and Lao, Shanshan and Gong, Yuan and Cao, Mingdeng and Wang, Jiahao and Yang, Yujiu},
  booktitle={Proceedings of the IEEE/CVF conference on computer vision and pattern recognition},
  pages={1191--1200},
  year={2022}
}

@inproceedings{ke2021musiq,
  title={Musiq: Multi-scale image quality transformer},
  author={Ke, Junjie and Wang, Qifei and Wang, Yilin and Milanfar, Peyman and Yang, Feng},
  booktitle={Proceedings of the IEEE/CVF international conference on computer vision},
  pages={5148--5157},
  year={2021}
}

@inproceedings{wang2023exploring,
  title={Exploring clip for assessing the look and feel of images},
  author={Wang, Jianyi and Chan, Kelvin CK and Loy, Chen Change},
  booktitle={Proceedings of the AAAI conference on artificial intelligence},
  volume={37},
  number={2},
  pages={2555--2563},
  year={2023}
}

@article{dosovitskiy2020image,
  title={An image is worth 16x16 words: Transformers for image recognition at scale},
  author={Dosovitskiy, Alexey and Beyer, Lucas and Kolesnikov, Alexander and Weissenborn, Dirk and Zhai, Xiaohua and Unterthiner, Thomas and Dehghani, Mostafa and Minderer, Matthias and Heigold, Georg and Gelly, Sylvain and others},
  journal={arXiv preprint arXiv:2010.11929},
  year={2020}
}

@article{lin2025diffusion,
  title={Diffusion adversarial post-training for one-step video generation},
  author={Lin, Shanchuan and Xia, Xin and Ren, Yuxi and Yang, Ceyuan and Xiao, Xuefeng and Jiang, Lu},
  journal={arXiv preprint arXiv:2501.08316},
  year={2025}
}

@ArtifactSoftware{R,
    title = {R: A Language and Environment for Statistical Computing},
    author = {{R Core Team}},
    organization = {R Foundation for Statistical Computing},
    address = {Vienna, Austria},
    year = {2019},
    url = {https://www.R-project.org/},
}


\end{document}